\pdfoutput=1

\documentclass[11pt]{article}

\usepackage{EMNLP2023}
\usepackage{graphicx} 
\usepackage{algorithm}
\usepackage{algpseudocode}
\usepackage{amsmath}
\usepackage{amsfonts}
\usepackage{times}
\usepackage{latexsym}
\usepackage{bm}
\usepackage{booktabs} 
\usepackage[T1]{fontenc}
\usepackage{xspace}
\usepackage{enumitem}

\usepackage[utf8]{inputenc}

\usepackage{microtype}

\usepackage{inconsolata}

\newcommand{\our}{\texttt{PromptRE}\xspace}

\title{\our: Weakly-Supervised Document-Level Relation Extraction via Prompting-Based Data Programming}

\author{Chufan Gao$^1$, Xulin Fan$^1$, Jimeng Sun$^1$, Xuan Wang$^2$ \\ 
$^1$University of Illinois Urbana-Champaign $^2$Virginia Tech\\ 
\texttt{chufan2@illinois.edu, xuanw@vt.edu}
}

\begin{document}
\maketitle
\begin{abstract}
Relation extraction aims to classify the relationships between two entities into pre-defined categories. While previous research has mainly focused on sentence-level relation extraction, recent studies have expanded the scope to document-level relation extraction. Traditional relation extraction methods heavily rely on human-annotated training data, which is time-consuming and labor-intensive. To mitigate the need for manual annotation, recent weakly-supervised approaches have been developed for sentence-level relation extraction while limited work has been done on document-level relation extraction. Weakly-supervised document-level relation extraction faces significant challenges due to an imbalanced number "no relation" instances and the failure of directly probing pretrained large language models for document relation extraction. To address these challenges, we propose \our, a novel weakly-supervised document-level relation extraction method that combines prompting-based techniques with data programming. Furthermore, \our incorporates the label distribution and entity types as prior knowledge to improve the performance. By leveraging the strengths of both prompting and data programming, \our achieves improved performance in relation classification and effectively handles the "no relation" problem. Experimental results on ReDocRED, a benchmark dataset for document-level relation extraction, demonstrate the superiority of \our over baseline approaches.
\end{abstract}

\section{Introduction}
Relation extraction is a crucial task in natural language processing that aims to classify the relationships between two entities (e.g., \texttt{Pacific Fair} and \texttt{Queensland}) into pre-defined categories (e.g., \texttt{located in}).
It has various downstream applications such as question answering \cite{veena2017graphQA} and knowledge graph construction \cite{RE_for_IR}. 

\begin{figure}[t]
    \centering
    \includegraphics[width=0.5\textwidth]{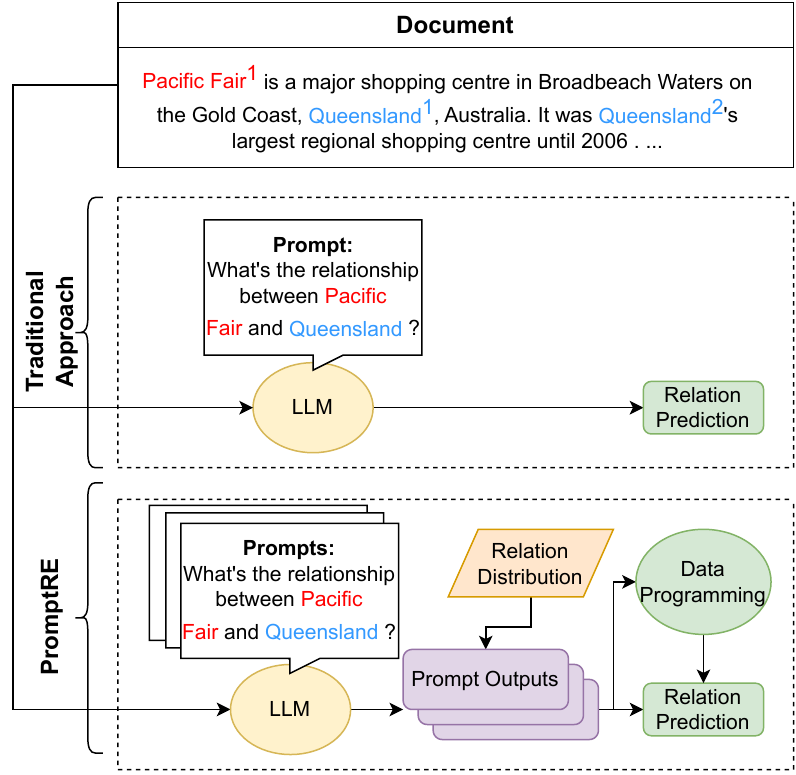}
    \caption{Differences between the naive approach and \our for weakly-supervised document-level relation extraction. We investigate various prompts and different ways to combine the prompting outputs using data programming. Furthermore, \our incorporates the entity type and relation distributions as prior knowledge to improve the classification performance. 
    }
    \label{fig:zerore_example}
\end{figure}

While previous research has mainly focused on relation extraction within a single sentence, recent studies have expanded the scope to document-level relation extraction \cite{yao2019docred}. Traditional relation extraction methods \cite{tan2022documentAdaptiveFocalLoss, ma2023dreeam} heavily rely on human annotation for training data, which is time-consuming and labor-intensive. To mitigate the need for manual annotation, recent weakly-supervised approaches \cite{sainz2021label, yang2023extracting} have been developed for relation extraction with minimal or no manual annotation. 
For example, \citet{qu2018weakly} extracted textual patterns from seed examples and use those patterns as weak supervisions for relation extraction. \citet{sainz2021label} represented each relation class using a label verbalizer and then solving the relation extraction task by a textual entailment model. \citet{wang2022unified} analyzed an "extremely unlabeled" scenario where each relation type had only one instance, reducing the training set to about five thousand labeled relation triplets. However, these methods were primarily designed and evaluated for sentence-level relation extraction, which limits their generalizability to document-level relation extraction datasets like ReDocRED \cite{tan2022revisiting}, where the presence of a substantial number of "No Relation" or NA classes poses additional challenges. 


To address this limitation, we study the problem of weakly-supervised document-level relation extraction. Recent large language models (LLMs) have achieved great success in a wide range of natural language processing tasks \cite{brown2020GPT, touvron2023llama}. 
We investigate the ability of the pretrained large language models on the document-level relation extraction task. We focus on three pretrained large language models: UnifiedQA \cite{2020unifiedqa, khashabi2022unifiedqa}, LlaMA,  LlaMA2 \cite{alpaca, touvron2023llama}, and ChatGPT \cite{ouyang2022training}.

\textbf{UnifiedQA} is a T5 model \cite{raffel2020exploring} pretrained on four different question answering settings: extractive, abstractive, multiple-choice, and yes/no questions. UnifiedQA performs comparably to specialized state-of-the-art models on most relation extraction datasets.
\textbf{ChatGPT}, developed by OpenAI, is a powerful generative large language model known for its impressive generalization capabilities. However, the closed-source nature of the ChatGPT model limits its accessibility for downstream applications. We utilize the text output of ChatGPT without accessing its internal embedding space or doing any model fine-tuning. 
 \textbf{LlaMA} is a collection of foundation language models ranging from 7B to 65B parameters trained on only publicly available datasets. 
After fine-tuning on an instruction-following dataset \cite{alpaca}, LlaMA and LlaMA2 are able to produce reasonable responses to the input instructions. We use LLaMA-7B and LLaMA2-7B, which has a good balance between model performance and efficiency.


We propose \our, a novel weakly-supervised document-level relation extraction method that combines prompting-based techniques with data programming (Figure~\ref{fig:zerore_example}). 
Given a known type-relation distribution, we first investigate various ways of prompting the pretrained large language models for relation classification. We then investigate different ways to select the most confident outputs using data programming, a technique that combines multiple sources of weak supervision. By leveraging the strengths of both prompting and data programming, we achieve improved performance in relation classification and effectively handle the "no relation" problem. Furthermore, we leverage ChatGPT as a summarizer to extract relevant information about the entities of interest. This plays a crucial role especially when dealing with lengthy documents which contains unnecessary extra information. To the best of our knowledge, we are the first to propose weakly-supervised document-level relation extraction. Our contributions are summarized as follows:
\begin{enumerate}[leftmargin=*]
    \item We propose the first weakly-supervised method, \our, for the document-level relation extraction task.
    \item \our is a novel method that combines various types of prompting outputs with data programming. \our further incorporates the label distribution and entity types as prior knowledge to improve performance.
    \item Extensive experiments on the ReDocRED dataset demonstrate the capability of \our over baseline methods. Ablations provide a comprehensive study on weakly-supervised inference ability. Multiple case studies show the incompleteness of existing document relation extraction datasets. 
\end{enumerate}

\section{Related Work}


\subsection{Document-level Relation Extraction}
Document-level relation extraction is a crucial task in natural language processing, as more than 40.7\% of relations requires multiple sentences to extract \cite{yao2019docred}. 
Consider an example of document-level relation extraction in Figure~\ref{fig:sample}. The task is to identify the relationship between a pair of entities ("Pacific Fair" and "Queensland") in the input document. Each of the entities has two mentions in the text (denoted by superscripts). To infer their relationship, it is evident that the mention-mention pair involving the first mention of each entity provides the most valuable information for extracting the relationship between them. 

Compared to sentence-level relation extraction, document-level relation extraction requires reasoning over multiple sentences which requires neural models to model long-range information. Additionally, entities may contain multiple mentions, which could include irrelevant information. However, this also allows for more information to model the relationship between entity-entity pairs. 

\begin{figure}[t]
  \texttt{{\color{red}Pacific Fair$^1$} is a major shopping centre in Broadbeach Water on the Gold Coast, {\color{blue}Queensland$^1$}, Australia. It was {\color{blue}Queensland$^2$}'s largest regional shopping centre until 2006. {\color{red}Pacific Fair$^2$} was developed by Hooker Retail Developments and opened in 1977 on what was swampland with 96 specialty stores and two anchor tenants.}
  \caption{Sample document relation extraction task from DocRED \cite{yao2019docred}. The {\color{red}red text indicates the head entity}, and the {\color{blue} blue text indicates the tail entity}. Here, head is related to tail by \texttt{"P131: located in the administrative territorial entity"}. 
  }
  \label{fig:sample}
\end{figure}

Pretrained language models, such as BERT-based models \cite{xu2021entityPLMBased}, have demonstrated significant success in document-level relation extraction. For example, BERT-based methods have employed techniques like hierarchical inference networks \cite{tang2020hin}, improved co-reference reasoning \cite{ye2020coreferential}, and adaptive thresholding. Additionally, graphical neural networks (GNNs) \cite{zeng2020doublegraph} have also been utilized for modeling document-level relation extraction. GNNs are used for feature learning on a coreference graph \cite{sahu2019inter}, edge-oriented learning techniques \cite{christopoulou2019connecting}, utilizing attention mechanisms \cite{guo2019attention}, and applying iterative refinement strategies for aggregating multi-hop information \cite{nan2020reasoning}. Moreover, several works have proposed new loss functions to tackle the class-imbalance problem in document-level relation extraction \cite{zhou2021documentadaptiveThre, tan2022documentAdaptiveFocalLoss}.

However, previous research on document-level relation extraction has relied heavily on human annotation for generating training data, which can be a time-consuming and labor-intensive process. Limited work has been conducted on document-level relation extraction methods that do not require human annotation.

\subsection{Weakly-Supervised Relation Extraction}
Weakly supervised methods have been extensively explored for relation extraction \cite{jiang2009multi, huang2017deep, qu2018weakly, wang2018open, li2018pattern}. For example, \citet{huang2017deep} utilized residual connections and convolutional neural networks (CNNs) to select relevant candidates to enhance supervised relation classification. \citet{qu2018weakly} extracted textual patterns from seed examples to provide additional supervision. \citet{phi2018ranking} introduced a ranking-based approach for seed selection, improving bootstrapping and distantly supervised relation extraction. \citet{sainz2021label} proposed representing each relation class using a label verbalizer and addressing the relation extraction task with a textual entailment model. \citet{wang2022unified} analyzed an "extremely unlabeled" scenario where each relation type had only one instance and reduced the training set to a smaller number of labeled relation triplets (but still contained more than 5000 training triplets). 

However, these methods were either primarily designed and evaluated for sentence-level relation extraction or still requires many labels, which limits their generalizability to our weakly-supervised document-level relation extraction task.

\section{Methodology}
\label{sec:Methodology}

\begin{figure*}[t]
    \centering
    \includegraphics[width=\textwidth]{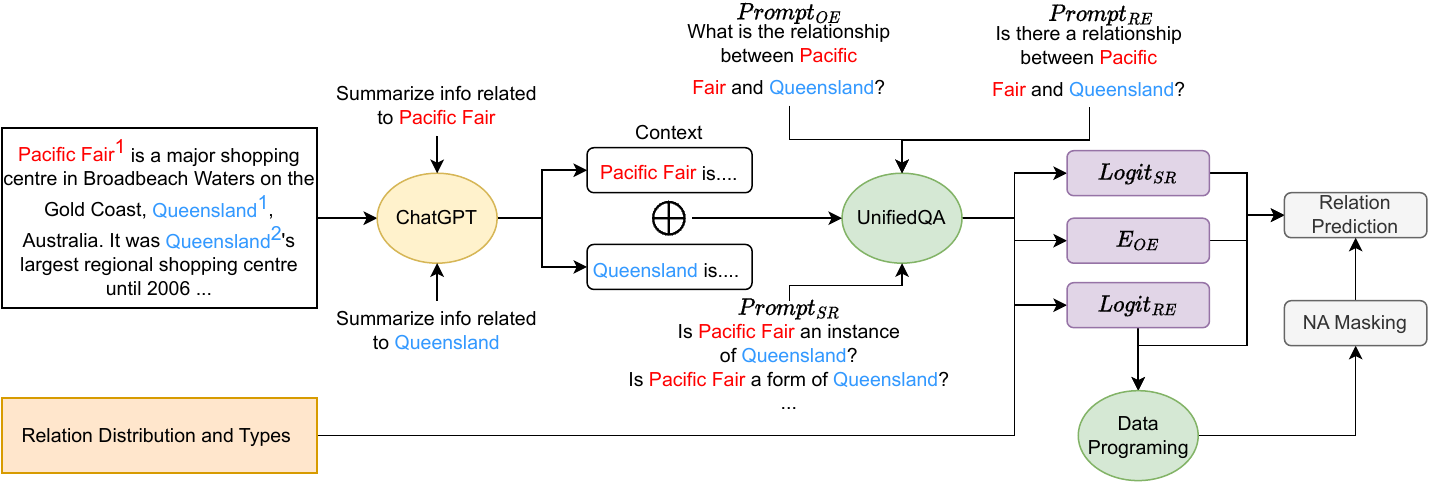}
    \caption{Overall framework of \our. Given an example document and a expected relation distribution, we first summarize the relevant portions of the text regarding both entities and concatenate ($\bm{\oplus}$) them together for entity-relevant context. Then, we use a variety of prompts to obtain: 1. The model prediction of each valid relation, 2. the open ended model relation prediction, and 3. The model prediction of the existence of a relationship. The three outputs are then used for relation prediction and data programming for addressing the "no relation" issue (referred to as NA Masking). 
    }
    \label{fig:zerore}
\end{figure*}

We propose \our, a weakly supervised document-level relation extraction method that combines large language model prompting with data programming. An illustration of the overall framework of \our is shown in Figure~\ref{fig:zerore}.


\subsection{Problem Definition}
In our task formulation, we consider a document $D$ consisting of $M$ sentences ($s_1$, $s_2$, ..., $s_M$) and $N$ entities ($e_1$, $e_2$, ..., $e_N$). Given this document $D$, a specified entity pair ($e_{head}$, $e_{tail}$), and a set of positive entity-entity relations ($r_1$, $r_2$, ..., $r_k$), the objective is to predict a set of relations ($\hat{r}_1, \hat{r}_2, ..., \hat{r}_p $) between the pair of entities based on the information extracted from the document. Note that each entity can have multiple occurrences within the document $D$.

\subsection{Entity-Oriented Document Preprocessing}
One challenge in document-level relation extraction is the long context. Models need to be able to find and focus on the information specific to the given pair of entities. In our \our pipeline, we leverage the power of ChatGPT to solve this problem. 
For example, given an entity  "Pacific Fair" and the document shown in Figure~\ref{fig:sample}, we ask ChatGPT "Based on the given paragraph, summarize the information about "Pacific Fair" \textbackslash n Pacific Fair is a major shopping center in Broadbeach Waters...". 

ChatGPT will generate a natural language summary of the information about the entity in the paragraph. We concatenate the summary of head and tail entities to form the text description of the two entities (denoted as \texttt{<context>}). In later stages, this summary is utilized as the context in place of the original document for relation prediction.

\subsection{Relation Prediction Via Prompting}
For predicting the relation class, we explore two approaches: Relation Specific Prompting and Open Ended Prompting. 
For both approaches, we query all non-identity entity-entity pairs in the documents. That is, if one document has $n_e$ entities, we query $n_e (n_e - 1)$ times.

\paragraph{Relation-Specific Prompting}
We prompt the large language models over all possible relation class, for all possible entity pairs. For each relation class, we hand-craft a yes-no question. 
For example, for the relation class "instance of", our hand-crafted version of the question is \texttt{"Is "Pacific Fair" an instance of "Queensland" <context> ?"}. 
To quantify the certainty of the large language model, we obtain a prediction score $Logit_{SR}$ by subtracting the logit of the 'no' output from the logit of the 'yes' output. This logit score $Logit_{SR}$ is calculated over each relation class and normalized to obtain a predicted relation class distribution.




\paragraph{Open-Ended Prompting} 
With the open-ended approach, we only prompt the large language models once for each entity pair with the question \texttt{"What's the relationship between "Pacific Fair" and "Queensland" <context> ?"}.
From there, we obtain the entity pair embedding as follows:
$$E_{OE} = \frac{\sum_{i=1}^{|embed|} LLM(P_{OE})_{embed}}{|embed|}$$ 
of the large language model output. Note that $E_{OE} \in \mathbb{R}^{N_{hidden}}$, where $N_{hidden}$ is of hidden dimension size from the large language model.\footnote{$N_{hidden}$ is 1024 for UnifiedQA and 4096 for LlaMA and LlaMA2.}

In addition, we encode each of the relation classes with a relation embedding $E_{rels} \in \mathbb{R}^{N_{rels} \times  N_{hidden}}$ where $N_{rels}$ is the total number of relation classes. We use cosine similarity between the entity pair embedding $E_{OE}$ and the relation embeddings $E_{rels}$ to compute a score over each relation, which is then normalized to obtain a predicted class relation distribution. 

\paragraph{Utilizing the Type Distribution} 
In our problem setup, we assume that we know the distribution of the relation classes given the types. We argue that this assumption, while strong, is reasonable. For example, it makes sense that no "Person - Person" entity pair could have the relation of "\texttt{country of citizenship}", as it does not make logical sense. It would be much more reasonable if the entity types were "location - person". We assume that this implicit knowledge is provided by the expert on the domain on which this framework is applied, and therefore we add the relevant relation distribution to the predicted probabilities of the raw scores, given the entity type pairs. In our experiments, we estimate this relation distribution from the ReDocRED dataset. More details can be found in Appendix~\ref{app:hyperparameters}.

\paragraph{Multi-Label Prediction}
Since each document has multiple possible labels, we take only the top $p$ percentile of confident predictions over all of the valid relation classes. Note that in this step, we do not consider the No-Relation class. 

\subsection{Addressing the No-Relation Issue}
Although we can extract potential relations via relation prompting from the previous section, we face the issue of false positives in relation prediction due to the large number of No-Relation classes. To address this, we design \our to choose only the most confident relation predictions.

\paragraph{Relation Existence Prompting} 
To obtain the model prediction for the existence of a relation in the input text, we prompt the model with the following prompt: \texttt{"Is there a relationship between "Pacific Fair" and "Queensland" <context> ?"}
To quantify the certainty of the large language model, we obtain a prediction score $Logit_{RE}$ by subtracting the logit of the 'no' output from the logit of the 'yes' output. This score is used to preserve only the most confident model predictions.

\paragraph{Data Programming (DP)} 
We combine multiple sources of weak supervision to select highly confident prediction from the previous step of relation prompting. Data programming is a framework to create denoised pseudo-labels from multiple sources of weak supervision from labeling functions \cite{ratner2016data, ratner2019training}.

A labeling function (LF) is a noisy heuristic that takes in data and assigns labels to unlabelled data or abstains from making a prediction. For example,  \texttt{f(text) = return SPAM if "http" in text else ABSTAIN} is a labeling function for spam detection.
At a high level, we frame the problem as dependency graph $G_{source}$ where each labeling function $\lambda_i$ is dependently conditioned on the true label $Y$. In our case, we assume conditional independence of all $\lambda_i | Y$. For this case, the dependency graphs will have observable cliques  $\bm{O} = \{\lambda_i, i \in n_{lf}\} \subset C$, where $n_{lf}$ is the number of labeling functions.

From here, we can analyze the covariance matrix of an observable subset of the cliques in $G_{source}$, leading to a matrix completion approach for recovering estimated accuracies $\mu$ (used in the final label model to predict $P(\bm{Y}|\bm{\lambda})$).

We assume that $\mu = \mathbb{E}(\psi(C))$ where $\psi(C)$ is vector of indicator random variables for all
combinations of all but one of the labels emitted by each variable in clique C.

The norm of the covariance of observed LFs cliques $O$ and separator set $S$ cliques $\bm{Cov}(\psi(O) \cup \psi(S))$ can be used to recover $\mu$. 

\begin{align}
& \bm{Cov}(\psi(O) \cup \psi(S)) = \Sigma =  \begin{bmatrix}
\Sigma_{O} & \Sigma_{OS} \\
\Sigma_{OS}^T & \Sigma_{S}
\end{bmatrix}
\end{align}
Its inverse is:
\begin{align}
& K = \Sigma^{-1} =  \begin{bmatrix}
K_{O} & K_{OS} \\
K_{OS}^T & K_{S}
\end{bmatrix}
\end{align}
Applying block matrix inversion, we get:
$$K_O = \Sigma^{-1}_O + c \Sigma^{-1}_O \Sigma_{OS} \Sigma^T_{OS} \Sigma^{-1}_{O}$$
$$c = (\Sigma_{S} - \Sigma^T_{OS} \Sigma^{-1}_{O} \Sigma_{OS})$$
Let $z = \sqrt{z} \Sigma^{-1}_{O} \Sigma_{OS}$, then
$$K_O = \Sigma^{-1}_O + zz^T$$
Solving for $z$ can directly recover $\mu$ via Algorithm 1 in \citet{ratner2019training}.

\paragraph{Reducing NA Predictions via Weak Supervision}
To address the "No relation" issue, we attempt to combine multiple sources of weak supervision through data programming to obtain a stronger prediction.
We consider three sources of weak supervision below.

The first source is the logit of relation-existence prompting $Logit_{RE}$. A higher logit indicates a better likelihood of a relationship between the pair of entities. Additionally, by rephrasing the prompt in different ways, we obtain different views of the model opinion on the existence of a relationship. Other paraphrases could include \texttt{"Is there a direct relationship between $e_{head}$ and $e_{tail}$?", "Does $e_{head}$ have any connection to $e_{tail}$?"}, and more.

The second source is the average logit of relation-specific prompting $Logit_{SR}$. The motivation is that if the entity pair has a low average logit for every relation-specific prompt, then it is not relevant to any of the relation classes and there is likely no relationship between the pair of entities. 

The second source is the average cosine similarity between the entity pair embedding $E_{OE}$ and the relation embedding $E_{Rels}$. Similar to the previous motivation for relation-specific prompting, if an entity pair embedding is very dissimilar from every relation embeddings, then there is likely no relationship between the pair of entities.

To summarize, we combine the three sources of weak supervision as input to the data programming model. Then, we take the $argmax$ from the probabilistic predictions of the data programming model and it as a mask to ensure that only the most probable predictions remain. Following the approach of \citet{ratner2019training}, we also fit a logistic regression model on $X=E_{OE}$ and label model predictions $\hat{Y}\sim P(\bm{Y}|\bm{\lambda})$ in order to smooth the decision boundaries. 

\begin{table}[t]
\centering
\caption{Statistics of the Re-DocRED dataset as well as the entire paraphrased ChatGPT Summary for every unique entity pair. Although the total number of unique documents is large, they are constructed by concatenating relevant information regarding both entities (and only require $n_e$ calls per document).}
\label{tab:data_stats}
\resizebox{.5\textwidth}{!}{%
\begin{tabular}{ccccc}
\toprule
 & \multicolumn{3}{c}{Re-DocRED} & \begin{tabular}[c]{@{}c@{}}ChatGPT \\ Summary\end{tabular} \\ \hline 
Stats                  & Train & Dev  & Test & Dev       \\ \hline
\# Docs           & 3,053 & 500  & 500  & 1,193,092 \\
Avg. \# Entities  & 19.4  & 19.4 & 19.6 & 19.4      \\
Avg. \# Triples   & 28.1  & 34.6 & 34.9 & 34.6      \\
Avg. \# Sentences & 7.9   & 8.2  & 7.9  & 5.3       \\ \bottomrule
\end{tabular}}
\end{table}

\begin{table*}[t]
\centering
\caption{We compare all results as ran on UnifiedQA-large, UnifiedQA-3b, and LlaMA-7b denoted by \{$large, 3b, llama, llama2$\} for different models. Simple RE denotes using the thresholded output of $Logit_{RE}$ without data programming. MV denotes using the baseline majority vote label model. DP denotes using data programming for weak supervision. Knowing the True NA Mask indicates using the ground truth relation existence labels. Bold denotes best performance.}
\label{tab:results}
\resizebox{.55\textwidth}{!}{%
\begin{tabular}{ccccc}
\toprule
            \textbf{Methods}                        & \textbf{F1}      & \textbf{Ign F1}  & \textbf{Precision}       & \textbf{Recall}       \\ \bottomrule
\textbf{Weakly Supervised Methods}  &         &         &         &         \\ \hline
Logits$_{large}$ + Simple RE        & 5.5975  & 4.8830  & 3.4246  & 15.3147 \\
Embed$_{large}$ Sim. + Simple RE    & 9.2030  & 7.9314  & 5.6304  & 25.1794 \\
Embed$_{large}$ Sim. + MV           & 9.5576  & 8.5099  & 7.9800  & 11.9128 \\
Embed$_{large}$ Sim. + DP (\our)           & 10.2232 & 8.7384  & 6.3969  & 25.4397 \\
\hline
Embed$_{3b}$ Sim. + Simple RE       & 9.1290  & 7.8723  & 5.5852  & 24.9769 \\
Embed$_{3b}$ Sim. + MV              & 9.4973  & 8.4576  & 7.9296  & 11.8375 \\
Embed$_{3b}$ Sim. + DP (\our)              & 10.1465 & 8.6738  & 6.3489  & 25.2488 \\
\hline
Embed$_{llama}$ Sim. + Simple RE    & 9.2136  & 7.9386  & 5.6369  & 25.2083 \\
Embed$_{llama}$ Sim. + MV           & 6.6330  & 6.0858  & 7.7816  & 5.7799  \\
Embed$_{llama}$ Sim. + DP (\our)          & 9.9368  & 8.5486  & 6.4909  & 21.1814 \\
\hline
Embed$_{llama2}$ Sim. + Simple RE    & 9.3214  & 8.0442  & 5.7029  & 25.5034 \\
Embed$_{llama2}$ Sim. + MV           & 8.1840  & 7.4837  & \textbf{8.4589} & 7.9264  \\
Embed$_{llama2}$ Sim. + DP (\our)          & \textbf{10.5586} & \textbf{9.0371}  & 6.5623  & \textbf{27.0019} \\\hline
\textbf{Knowing the True NA Mask}         &         &         &         &         \\ \hline
Embed$_{large}$ Sim. + True NA Mask & 46.6324 & 42.1369 & 38.3962 & 59.3670 \\
Embed$_{3b}$ Sim. + True NA Mask    & 46.4416 & 41.9580 & 38.2390 & 59.1240 \\
Embed$_{llama}$ Sim. + True NA Mask & 46.6915 & 42.1909 & 38.4448 & 59.4423 \\ 
Embed$_{llama2}$ Sim. + True NA Mask & 46.7824 & 42.2882 & 38.5197 & 59.5580 \\ \hline
\textbf{Supervised Methods}         &         &         &         &         \\ \hline
DREEAM \cite{ma2023dreeam}                               & 80.73   & 79.66   & -       & -       \\
KD-DocRE \cite{tan2022documentAdaptiveFocalLoss}                            & 78.28   & 77.60   & -       & -    \\  
\bottomrule
\end{tabular}
}
\end{table*}

\begin{table}[t]
\centering
\caption{Experimental results with or without using ChatGPT for entity-oriented document preprocessing. $large$ and $3b$ denote the UnifiedQA model we use to compute cosine similarities.}
\label{tab:ablation3}
\resizebox{.5\textwidth}{!}{%
\begin{tabular}{ccccc} \toprule
\textbf{\textit{w/o} ChatGPT}                 & \textbf{F1}     & \textbf{Ign F1} & \textbf{Precision}      & \textbf{Recall}       \\ \hline
$_{large}$ + Simple RE       & 9.1924 & 7.9171 & 5.6240 & 25.1504 \\
$_{large}$ + DP              & 9.8235 & 8.6655 & \textbf{6.6853} & 18.5142 \\ \hline
$_{3b}$ + Simple RE       & 9.1861 & 7.9125 & 5.6201 & 25.1331 \\
$_{3b}$ + DP              & 9.9087 & 8.7183 & 6.5299 & 20.5334
\\ \hline
\textbf{\textit{w/} ChatGPT}                 & & & & \\ \hline
$_{large}$ + DP           & \textbf{10.2232} & \textbf{8.7384}  & 6.3969  & \textbf{25.4397}
\\ \bottomrule
\end{tabular}
}
\end{table}

\begin{table}[t]
\centering
\caption{Experimental results with Relation Type Distribution using Logits$_{large}$ as the baseline model.}
\label{tab:ablation2}
\resizebox{.5\textwidth}{!}{%
\begin{tabular}{ccccc} \toprule
\textbf{No Type Dist.}    & \textbf{F1}     & \textbf{Ign F1} & \textbf{Precision}      & \textbf{Recall}       \\ \hline
Simple RE         & 0.3997 & 0.3609 & 0.2445 & 1.0935  \\
DP                & 0.3604 & 0.3332 & 0.2278 & 0.8621  \\ \hline
\textbf{Only Type Dist.} &        &        &        &         \\ \hline
Simple RE         & 3.4499 & 2.8987 & 1.8514 & 25.2430 \\
DP                & 4.6606 & 3.8770 & 2.5747 & 24.5429 \\ \bottomrule
\end{tabular}
}
\end{table}

\begin{table*}[t]
\centering
\small
\caption{An example of a relation prediction of a portion of  labeled positive relation tuples. Each relation tuple has the format: (Head entity, Tail entity, Ground Truth Relation). The predicted text is shown below after prompting the model using an open ended prompt (e.g. "Is there a relationship between $e_{head}$ and $e_{tail}$?"). }
\label{tab:case_study_1}
\begin{tabular}{p{\textwidth}}
\toprule
\textit{Labeled} Relation Tuple: (Ross Patterson Alger; August 20 , 1920; date of birth) \\
{\fontfamily{pcr}\selectfont \small
Ross Patterson Alger was born on August 20, 1920 in Prelate, Saskatchewan. He became a politician and served as mayor of Calgary from 1977 to 1980.
} \\ \hline
\textit{Labeled} Relation Tuple: (Ross Patterson Alger; January 16 , 1992; date of death) \\
{\fontfamily{pcr}\selectfont \small
The relationship between "Ross Patterson Alger" and "January 16 , 1992" is that Ross Patterson Alger died on January 16, 1992 due to cancer which was initially detected in 1985...
} \\ \hline
\textit{Labeled} Relation Tuple: (Canadian; Alberta; contains administrative territorial entity) \\
{\fontfamily{pcr}\selectfont \small
Yes, there is a relationship between "Canadian" and "Alberta". Ross Patterson Alger was born in Saskatchewan and moved to Alberta...
} \\ \hline
\textit{Unlabeled} Relation Tuple: (Ross Patterson Alger; 1942; no relation) \\
{\fontfamily{pcr}\selectfont \small
Ross Patterson Alger was born in Saskatchewan, received a bachelor of commerce degree from the University of Alberta in 1942, and died in 1992 after battling cancer. He served in the Royal Canadian Air Force during World War II.} \\ \hline
\textit{Unlabeled} Relation Tuple: (Ross Patterson Alger; Calgary City Council; no relation) \\
{\fontfamily{pcr}\selectfont \small
Ross Patterson Alger was an alderman on Calgary City Council from 1971 to 1974 before being elected as the mayor in 1977.
} \\ 
\bottomrule
\end{tabular}
\end{table*}

\section{Experiments}

\subsection{Dataset}
To evaluate our methodology, we use ReDocRED \cite{tan2022revisiting}, an open-access, document-level relation extraction dataset that improves upon the popular DocRED dataset \cite{yao2019docred} by resolving incompleteness, addressing logical inconsistencies, and correcting coreferential errors. 
Table~\ref{tab:data_stats} shows the amount of training data available for all data splits as well as the ChatGPT-paraphrased entity-relevant text summary. Note that we primarily use the Dev set of ReDocRED for our experiments for computational practicality. 

\subsection{Experimental Settings}
For the large language models for relation extraction, we compared UnifiedQA \cite{2020unifiedqa, khashabi2022unifiedqa} (both \textit{3b} and \textit{large} versions) and Alpaca-lora\footnote{\url{https://github.com/tloen/alpaca-lora}}--a reproduction of the Stanford Alpaca LlaMA model \cite{alpaca, touvron2023llama, wang2022self} using LoRa \cite{hu2021lora}. 

Some experiments can only be run with a subset of these models. For example, $logits_{SR}$ is highly expensive to compute as it requires $(n_{e}^2 - n_{e}) \times N_{rels}$, so we only run the UnifiedQA-Large for this score computation. For all other score computations, we may use all models: UnifiedQA-3b, UnifiedQA-3b, and LLama-7b as the base models. 
Additionally, we also perform weak supervision experiments without $logits_{SR}$ due to its high cost (See App.~\ref{app:logits_sr} for more details). 
In our experiments, we use precision, recall, and F1 scores as the evaluation metrics for the performance comparison. More details about these evaluation metrics can be found in Appendix~\ref{app:Evaluation_Metrics}.


\subsection{Results}
Table~\ref{tab:results} shows the main results of our experiments. 
We observed that the logit performance of prompting every entity-entity pair with the relevant relation prompt does not perform as well as using the cosine similarity of the open ended QA embeddings and the prompt embeddings.
We suspect that this may be due to a number of reasons, including the lack of regularization on the score output.
Additionally, it is possible that using cosine similarity allows the model to capture a more semantically meaning snapshot of its response, rather than just a single scalar value.

As expected, using the ground truth NA labels leads to a large improvement over relaxing the assumption. It demonstrates the difficulty in determining the existence of relations in documents under weakly supervision and points out an exciting direction for future research.

\paragraph{Effect of Language Model Size}
The performance comparison between different model sizes is shown in Table~\ref{tab:results}.
One obervation is that the UnifiedQA-large model performs better than the UnifiedQA-3b model across all the metrics. \citet{khashabi2022unifiedqa} observed similar results between the large and 3b models, so this is not entirely unexpected. Additionally, it is interesting to see that even the LLaMA-7b model, the largest model we consider, provides no significant benefit over using the UnifiedQA-large model. This implies that prompting by itself does not work well for document-level relation extraction, as we see that the relations predicted are generally not of the same format as the true relation classes (see Section~\ref{sec:Case_Studies}).

\paragraph{Effect of Entity-Oriented Document Preprocessing}
In this ablation study, We investigate the effects of using ChatGPT for entity-oriented document preprocessing (inspired by \cite{yu2022generate}). From Table~\ref{tab:ablation3}, we see that using the ChatGPT-generated entity-oriented document summaries yields better performance compared to using the original document as input for relation extraction. However, this suggests that further research could potentially avoid expensive calls to the private model without sacrificing too much performance.

\paragraph{Effect of Relation Type Distribution}
We further investigate the effect of relation type distribution on relation class prediction. As shown in Table~\ref{tab:ablation2}, the type distributions are strong signals as prior knowledge to enhance the performance of weakly-supervised relation extraction. We observe that using type distribution only without any other weak supervision achieves almost half of the performance of \our in Table~\ref{tab:results}. Our results from Table~\ref{tab:results} further show that the combination of prompting + type distribution performs the best. 

\subsection{Case Studies}
\label{sec:Case_Studies}

We analyze some example outputs from the predictions of the LLaMA-7b model, as shown in Table~\ref{tab:case_study_1}.
We see that practically, the LlaMA model output is biased towards much longer and more detailed text than is required for the relation prediction problem. This could explain why the embeddings between the answers and the relation text would be difficult to correlate, leading to worse performance. Furthermore, the third example shows an instance of an indirect relation. It is true that Ross Patterson Alger was born in Canada and moved to another part of Canada. This is a common failure case with the responses--the relation prediction is too specific to the original text. The final two examples indicate a weakness in the dataset. As with any relation extraction dataset, ReDocRED is not complete, and the large language model was able to pick up on two relations not in the ground truth labeled set--"received a bachelor's degree from" alderman" respectively. 

\section{Conclusion}
In this paper, we investigate several methods to integrate prompting and data programming for relation classification and evaluated our model on ReDocRED.
Results show that our best results yields around 10.2 F1 on the development set, a promising result for almost no supervision. Since this is a novel application, further research is required to investigate strategies for improvement. Some ideas include the following.

\textbf{The NA Issue:} The large number of "no relations" continues to be an issue for less than supervised methods for document relation extraction on existing datasets. Further work should focus on more efficient and accuracy ways to mine distant labels to address this issue. One major roadblock that coincides with NA is the lack of complete labels in the dataset as shown in the case studies. Future work could improve on existing document relation extraction datasets accordingly.

\textbf{Extending to the Few-Shot Case:} It is usually possible to query human experts for a few examples of the required classification task. Researching ways to take maximum advantage of a small set of labels would also be highly practical, and would not require much extra effort on the annotators. This could also tune a model to better address the "No Relation" issue.

\textbf{Final Thoughts:} We find that weakly supervised document-level relation extraction is a uniquely difficult problem due to the incomplete labels in popular datasets, and we propose \our to attempt to solve it via combining prompting and data programming. We show the effect of tuning different experimental setups, including model size, entity-oriented summarization, and the effect of our relation-type distribution assumption. Case studies support the finding that existing document-level relation extraction datasets may be severely lacking in label completeness. Although the results are a considerable margin from contemporary supervised methods, we hope that this work can serve as a stepping stone in this novel area of less-than-supervised document relation extraction.

\clearpage
\section*{Limitations}
Although we investigated multiple different LLMs and parameters and the type relation distribution for relation prediction as well as addressing the false positives, the performance we attained is still limited compared to supervised methods on the same task. 
Additionally, relation prediction is dependent on the prompt choice, as we see from the open-ended prompts performing better than asking specific relationships. Data Programming is also dependent on high quality sources of weak supervision, as we see from the improvement in performance when not considering the logits in Table~\ref{tab:ablation2}. 
Effectively mapping the output of language models to the concrete label space without training remains an hard problem for future work to tackle.

\section*{Ethical Statement}
Based on the methodology we have currently employed, we do not foresee any significant ethical concerns. 
All the documents and models utilized in our study were obtained from open-source domains, ensuring a transparent and accessible source of information. 
Additionally, \our requires no LLM training, eliminating the risk of model drift.
Additionally, the task of relation extraction is a widely recognized and well-studied problem across various natural language processing applications.

However, it is crucial to acknowledge a minor factor, namely the presence of potential hidden biases within the pretrained language models used in our analysis. 
These biases may stem from the data on which the models were trained, which could have inadvertently introduced implicit human biases. 
While our usage of these pretrained language models enable us to identify relationships between arbitrary entities, it is conceivable that biases may emerge if one were to explore sensitive relation classes and entities.

\bibliography{custom}
\bibliographystyle{acl_natbib}

\clearpage
\appendix

\section{Parameter Settings}
\label{app:hyperparameters}
All models were ran on an NVIDIA A6000 with 48 gigabytes of VRAM. Still, around 10 days were required to fully run the experiments. For particularly expensive computations, like $Logits_{SR}$, only the fastest model--UnifiedQA-large--could be feasibly ran.

All models were downloaded from Huggingface \cite{wolf2019huggingface}. We use the default setup of the pretrained models and did not do further finetuning. All the step mentioned in the methodology section works on the output of the pretrained models.

Supervised results DREEEAM \cite{ma2023dreeam} and KD-DocRE \cite{tan2022documentAdaptiveFocalLoss} were taken from the original source papers. 

\section{Evaluation Metrics}
\label{app:Evaluation_Metrics}
To keep in tradition with existing document relation extraction work, we report both F1 and Ign\_F1 as computed by the official metrics from ReDocRED. 
F1 refers to micro-averaged F1 score that combines precision $P$ and recall $R$ $$F1 = \frac{2 P R}{P+R}$$
$$P = \frac{\texttt{length of correct (h,t,rel) preds}}{\texttt{length of all  (h,t,rel) preds}}$$
$$R = \frac{\texttt{length of correct (h,t,rel) preds}}{\texttt{length of correct  (h,t,rel)}}$$
Where \texttt{(h,t,rel)} denotes a tuple of the predicted head, tail, and relation.
Ign\_F1 is computed similarly to above, but ignores the samples in the DocRED's distantly supervised training set. (Note that we do not use any distantly labelled data).


\section{Effect of Relation-Specific Prompts}
\label{app:logits_sr}
In this ablation study, we investigate the usefulness of $Logits_{SR}$ on both the relation class prediction part as well as the addressing the NA issue. Because each entity-entity pair has to prompt with all relations, it is quite expensive to perform. Thus, we only perform experiments with the fastest model we consider--UnifiedQA large. From Table~\ref{tab:results}, we see that by itself, using the logits do not perform as well as embedding similarity. 

For the case of reducing NAs, we actually do not include it as a source of weak supervision in the data programming framework due to its inefficiency. However, if we \textit{did} include it, we would see that performance drops as well, as shown in Table~\ref{tab:ablation1}.

\begin{table}[t]
\centering
\caption{Experimental results via prompting the model for each specific relation using the baseline model Logits$_{large}$.}
\label{tab:ablation1}
\resizebox{.5\textwidth}{!}{%
\begin{tabular}{ccccc}
\toprule 
 & \textbf{F1}      & \textbf{Ign F1}  & \textbf{Precision}       & \textbf{Recall}      \\ \hline
Simple RE    & 5.5975  & 4.8830  & 3.4246  & 15.3147 \\
MV      & 4.5163  & 4.3027  & 6.6494  & 3.4193  \\
DP           & 6.2096  & 5.4219  & 3.9254  & 14.8519 \\
\bottomrule
\end{tabular}
}
\end{table}

\begin{figure}[t]
  \texttt{\small Ross Patterson Alger ( August 20 , 1920 – January 16 , 1992 ) was a politician in the Canadian province of Alberta , who served as mayor of Calgary from 1977 to 1980 . Born in Prelate , Saskatchewan , he moved to Alberta with his family in 1930s . He received a bachelor of commerce degree from the University of Alberta in 1942 . He served with the Royal Canadian Air Force during World War II . After the war , he received an MBA from the University of Toronto . He settled in Calgary and started a career in accounting . In 1958 , he was a public school board trustee , and later became the chairman . From 1971 to 1974 , he was an alderman on Calgary City Council . In 1974 , he ran for mayor losing to Rod Sykes . He was elected mayor in 1977 and served one term until 1980 . During Alger 's term , notable accomplishments include the construction of the Ctrain ’s first leg , the bid for the XV Olympic Winter Games , and planning for the Olympic coliseum . His brother was Harry Alger . Alger died of cancer in 1992 , which had first been diagnosed in 1985 .
}
  \caption{Original document for Case Study 1.}
  \label{fig:case_study_1}
\end{figure}

\begin{figure}[t]
  \texttt{\small  Mess of Blues is an album by Jeff Healey . It was released in 2008 less than two weeks after his death and just three weeks shy of his 42nd birthday . Four of the album 's tracks were recorded live in front of audiences , two of the live tracks at the Islington Academy in London , and the other two live tracks at Jeff Healey 's Roadhouse in Toronto . The other six tracks were recorded at Studio 92 in Canada by Norm Barker and Richard Uglow . The whole album features the band which normally accompanied Jeff at his club , Jeff Healey 's Roadhouse . The song " Mess of Blues " , which appears on the album was written by Doc Pomus and Mort Shuman and was originally recorded by Elvis Presley .
}
  \caption{Original document for Case Study 2.}
  \label{fig:case_study_2}
\end{figure}

\section{Relation Distribution Calculation}
\label{app:Relation_Distribution_Calculation}
We test our assumptions of the relation type distribution. Specifically, we how the performance changes with more or less expert annotated documents. The results are shown in Table~\ref{tab:ablation4}. Recall that in total, we have 500 documents, so 1\% of all documents represents only 5 annotated documents. This reinforces our assumption that creating this relation/type distribution is not exorbitantly expensive. Furthermore, this computation is only an \textit{estimate} of the actual input that domain experts would provide, so it is possible that real world performance would be better or worse depending on the distribution of true types and relations.

\begin{table*}[t]
\centering
\caption{The Performance of UnifiedQA-large on varying percentage of the data we use for to compute the expert-provided relation/type distribution.}
\label{tab:ablation4}
\begin{tabular}{ccccc}
\toprule
1\%                              & F1      & Ign F1 & Precision & Recall  \\ \hline
Embed$_{large}$ Sim. + Simple RE & 6.8663  & 5.8224 & 4.2008    & 18.7862 \\
Embed$_{large}$ Sim. + DP        & 7.8207  & 6.5522 & 4.9439    & 18.7052 \\ \hline
10\%                             &         &        &           &         \\ \hline
Embed$_{large}$ Sim. + Simple RE & 8.6363  & 7.4106 & 5.2837    & 23.6288 \\
Embed$_{large}$ Sim. + DP        & 9.6495  & 8.1978 & 6.0999    & 23.0791 \\ \hline
25\%                             &         &        &           &         \\ \hline
Embed$_{large}$ Sim. + Simple RE & 8.9535  & 7.7211 & 5.4778    & 24.4966 \\
Embed$_{large}$ Sim. + DP        & 9.8986  & 8.4370 & 6.2575    & 23.6751 \\ \hline
50\%                             &         &        &           &         \\ \hline
Embed$_{large}$ Sim. + Simple RE & 9.1121  & 7.8481 & 5.5748    & 24.9306 \\
Embed$_{large}$ Sim. + DP        & 9.9107  & 8.4176 & 6.2651    & 23.7040 \\ \hline
Original (100\%)                 &         &        &           &         \\ \hline
Embed$_{large}$ Sim. + Simple RE & 9.2030  & 7.9314 & 5.6304    & 25.1794 \\
Embed$_{large}$ Sim. + DP        & 10.2232 & 8.7384 & 6.3969    & 25.4397 \\
\bottomrule
\end{tabular}
\end{table*}

\section{Additional Case Studies}
\label{app:More_Case_Studies}
We analyze a few more examples of the model output in Table~\ref{tab:case_study_2} with the original document in Figure~\ref{fig:case_study_2}. While the first 3 labeled examples make intuitive sense and are correct, the lack of label completeness evident in the latter 3 examples. For example, in the last example, Richard Uglow is the recorder of part of the album, but the pair "Richard Uglow" and "Mess of Blues" is marked as "No relation" in the dataset. This type of weakly labeled situation is a particularly difficult problem for weakly-supervised methods like ours because without training, it's very hard to inform the model with subtle details regarding the relation label space. 

\begin{table*}[t]
\centering
\caption{An example of a relation prediction of a portion of  labeled positive relation tuples. Each relation tuple has the format: (Head entity, Tail entity, Ground Truth Relation). The predicted text is shown below after prompting the model using an open ended prompt (e.g. "Is there a relationship between <Head> and <Tail>?").}
\label{tab:case_study_2}
\begin{tabular}{p{\textwidth}}
\toprule
\textit{Labeled} Relation Tuple: (Mess of Blues; 2008; publication date) \\
{\fontfamily{pcr}\selectfont \small
The album Mess of Blues was released in 2008, less than two weeks after Jeff Healey's death and just three weeks shy of his 42nd birthday. It features four live tracks recorded in London and Toronto, and six studio tracks recorded in Canada.
} \\ \hline
\textit{Labeled} Relation Tuple: (Mess of Blues; Doc Pomus; lyrics by) \\
{\fontfamily{pcr}\selectfont \small
The song "Mess of Blues" was originally recorded by Elvis Presley and appears on Jeff Healey's posthumous album. Doc Pomus is mentioned as a songwriter who co-wrote the song with Mort Shuman.
} \\ \hline
\textit{Labeled} Relation Tuple: (Elvis Presley; Mess of Blues; notable work) \\
{\fontfamily{pcr}\selectfont \small
 song "Mess of Blues" on the album was originally recorded by Elvis Presley.
} \\ \hline
\textit{Unlabeled} Relation Tuple: (Mess of Blues; two weeks; no relation) \\
{\fontfamily{pcr}\selectfont \small
The album Mess of Blues by Jeff Healey was released two weeks after his death. He died just three weeks shy of his 42nd birthday.
} \\ \hline
\textit{Unlabeled} Relation Tuple: (Mess of Blues; Islington Academy; no relation) \\
{\fontfamily{pcr}\selectfont \small
The album "Mess of Blues" was recorded at the Islington Academy in London.
} \\ \hline
\textit{Unlabeled} Relation Tuple: (Mess of Blues; Richard Uglow; no relation) \\
{\fontfamily{pcr}\selectfont \small
Yes, there is a relationship between "Mess of Blues" and "Richard Uglow". Richard Uglow recorded six tracks of Jeff Healey's album "Mess of Blues" at Studio 92 in Canada along with Norm Barker.
} \\ 
\bottomrule
\end{tabular}
\end{table*}
\end{document}